\documentclass[11pt]{article}

\usepackage[final]{acl}

\usepackage{times}
\usepackage{latexsym}
\usepackage[T1]{fontenc}
\usepackage[utf8]{inputenc}
\usepackage{microtype}
\usepackage{inconsolata}
\usepackage{graphicx}
\usepackage{booktabs}
\usepackage{amsmath}
\usepackage{url}

\title{BioPhys-Bridge: A Benchmark for Interdisciplinary Scientific Reasoning in Physics-Grounded Biological Research}

\author{Qingyang Xu \\
  Independent Researcher \\
  Shanghai, China \\
  \texttt{qyxu1994@gmail.com}}

\begin{document}

\maketitle

\begin{abstract}
Language models face unique challenges in analyzing interdisciplinary scientific research literature. In biophysics research, faithful answers require grounding observed data in source evidence, interpreting it through a quantitative physics model, and linking it to a biological mechanism. To address this challenge, we introduce BioPhys-Bridge, a novel benchmark dataset for evidence-grounded scientific reasoning over biophysical literature. Each case contains evidence blocks, stable evidence IDs, quantitative values, units, equations, assumptions, mechanisms, and next decisions as grounding targets for question answering (QA) and retrieval-augmented generation (RAG). The initial release contains 500 cases, 1,517 agent-facing tasks, and covers six biological domains and nine physical model families, including three sparse families reserved for future expansion. We enforce strict quality gates for all cases in schema, evidence-integrity, quantitative-grounding, source-license, duplicate, unit-normalization, with domain expert review and annotation for 81 cases. Preliminary evaluations show that DeepSeek-V4-Flash obtain the highest evidence-ID $F_1$ score (0.360), followed by Qwen3.7-Max (0.316) and GPT-4o-mini (0.294). BioPhys-Bridge is an interdisciplinary benchmark for evaluating attribution, faithfulness, hallucination reduction, and biological experiment design with complex, multi-step scientific reasoning. Future works will increase the size and complexity of the dataset and perform comprehensive evaluations. Code and data are available in the \href{https://github.com/qyxu1994/BioPhys-Bridge}{GitHub repository} and on \href{https://huggingface.co/datasets/qyxu1994/BioPhys-Bridge}{Hugging Face}.
\end{abstract}

\section{Introduction}

Language models (LM) are increasingly used to analyze scientific research papers in retrieval-augmented generation (RAG) systems and other agentic AI frameworks. However, retrieving a relevant paragraph does not guarantee the faithfulness of the final answer: the AI agent can cite loosely related evidence, omit key assumptions, mix units, or
produce a plausible scientific interpretation that the cited paper does not support. Scientific grounding requires finding faithful connections among source evidence, quantitative measurements, model assumptions, and the leveraging domain expertise to derive scientific conclusions and hypotheses.

Biophysics is an epitome of this challenge due to its interdisciplinary nature. The LM needs to both perform semantic text matching and draw quantitative predictions from a biophysical model, in order to connect a quantitative prediction (e.g., binding affinity, reaction rate, phase-separation threshold) to the model assumptions under which it is meaningful, explain the underlying biological mechanism, and suggest the next experiment or computation. Existing scientific QA and RAG benchmarks often focus on gauging whether an answer is supported by textual evidence. In contrast, BioPhys-Bridge gauges whether LM can perform evidence-grounded reasoning across modalities (equations and semantics) and scientific disciplines.

We propose a novel framework for grounding LM which requires LM to
identify source evidence, use quantitative values and units consistently,
respect physics model assumptions, and produce an actionable response supported by evidence IDs. Evidence blocks are the
atomic citation units, while equations, assumptions, quantitative evidence, and
mechanism annotations provide intermediate structure for auditing whether a
model's answer is faithful to the research question and biological reasoning rather than superficial semantic relations.

The initial release of BioPhys-Bridge contains 500 cases extracted from open-access peer-reviewed research papers in biophysics. This release contains 1,517 distinct tasks for biological research. Every case passed rigorous schema validation,
evidence-integrity checks, quantitative-grounding checks, source-license
coverage checks, unit normalization, duplicate checks, and semantic content
filters. To validate the correctness and scientific value of the curated samples, we asked domain experts to review and annotate 81 cases, including the full held-out test set (with 50 cases) and additional golden examples.

This works makes five important contributions for grounded scientific LM.
First, we introduce a novel benchmark dataset for literature-grounded scientific
reasoning over biophysical papers. Second, we define a structured case schema
that treats evidence blocks, quantitative values, equations, assumptions,
mechanisms, and next decisions as first-class grounding objects. Third, we
provide a task suite for derivation, mechanism-from-evidence, discrepancy
explanation, and next-experiment design, each requiring supporting evidence
IDs. Fourth, we describe a validated release pipeline with schema,
evidence-integrity, quantitative-grounding, license, unit-normalization,
duplicate, and content-quality gates. Finally, we provide a de-leaked evaluation
harness with preliminary baselines that separate deterministic evidence
retrieval from answer generation on the same held-out tasks.

As far as we know, BioPhys-Bridge is the first benchmark for grounding LMs in
interdisciplinary scientific reasoning in physics-grounded biological research, where LM must connect evidence IDs to physics equations, model assumptions, biological mechanisms, and guide experiment design. It is useful for studying faithfulness, attribution, and hallucination in real-world research scenarios.

\section{Related Work}

\subsection{Grounded Language Models and Scientific RAG}

Scientific RAG systems aim to answer research questions by retrieving and
synthesizing evidence from papers. PaperQA and LitQA study
retrieval-augmented question answering over scientific literature
\citep{lala2023paperqa}. LitSearch evaluates realistic scientific literature
search queries over recent papers \citep{ajith2024litsearch}. LAB-Bench
measures biology research capabilities, including literature reasoning,
database use, figure interpretation, and sequence manipulation
\citep{laurent2024labbench}. SPIQA targets multimodal question answering over
scientific paper figures \citep{pramanick2024spiqa}. These resources make
scientific documents central to evaluation, but their primary supervision is
usually an answer, paper-level evidence, retrieval target, or document QA
context. BioPhys-Bridge instead makes evidence blocks, evidence IDs,
quantitative values, units, equations, model assumptions, and mechanism links
explicit fields in each case, so attribution can be evaluated beyond passage
retrieval.

BioPhys-Bridge also builds on scientific document parsing and evidence
extraction. MinerU provides open-source document content extraction for
complex PDFs, including layout, table, formula, and OCR processing
\citep{wang2024mineru}. In our pipeline, MinerU is the parsing layer from
which normalized evidence blocks are derived, audited, and linked to
structured scientific reasoning fields.

\subsection{Scientific Reasoning and Multimodal Benchmarks}

Scientific reasoning benchmarks increasingly test difficult science questions.
ScienceQA introduced a multimodal science question-answering benchmark with
explanations \citep{lu2022scienceqa}. SciBench evaluates college-level
scientific problem solving across mathematics, chemistry, and physics
\citep{wang2023scibench}. GPQA uses expert-written graduate-level biology,
physics, and chemistry questions to stress scalable oversight
\citep{rein2023gpqa}. MathVista and MMMU combine visual understanding with
expert or college-level reasoning \citep{lu2024mathvista,yue2024mmmu}, while
LongBench tests long-context understanding over extended inputs
\citep{bai2024longbench}. These benchmarks are valuable tests of scientific
knowledge, reasoning, and multimodal understanding, but they generally do not
represent a source-paper case as an evidence-linked chain from measurement to
physical model to biological mechanism and scientific decision.

\subsection{AI-for-Science and Physical-Model Benchmarks}

Scientific machine-learning resources such as PDEBench and BubbleML provide
physics-grounded data, benchmark tasks, and baselines for simulation and
surrogate modeling \citep{takamoto2022pdebench,jakli2023bubbleml}. Other
realistic agent benchmarks, including SWE-bench, WebArena, AgentBench, MINT,
ToolBench, and MLE-bench, evaluate software repair, web navigation, tool use,
and machine-learning engineering with functional, human-referenced, or
environment-based scoring
\citep{jimenez2024swebench,zhou2024webarena,liu2024agentbench,wang2024mint,qin2024toolllm,shern2025mlebench}.
BioPhys-Bridge is complementary: it does not benchmark a PDE solver, surrogate
model, or general-purpose tool agent directly. Instead, it evaluates
source-paper grounding for quantitative and mechanistic scientific
interpretation. BioPhys-Bridge sits
between literature-grounded QA and AI-for-science benchmarks: it evaluates
whether grounded language models can use source evidence to support
quantitatively and mechanistically meaningful scientific interpretations.

\begin{table*}[t]
\centering
\small
\begin{tabular}{p{0.18\linewidth}p{0.18\linewidth}p{0.25\linewidth}p{0.29\linewidth}}
\toprule
Benchmark family & Core unit & Primary strength & Gap addressed by BioPhys-Bridge \\
\midrule
PaperQA, LitQA \citep{lala2023paperqa}; LAB-Bench \citep{laurent2024labbench} & Scientific papers and research questions &
Literature-grounded QA, biology research skills, RAG evaluation &
Equations, units, assumptions, and mechanism links are not first-class case fields. \\
\midrule
ScienceQA \citep{lu2022scienceqa}; SciBench \citep{wang2023scibench}; GPQA \citep{rein2023gpqa} & Science questions or problems &
Scientific knowledge, explanations, expert difficulty &
No explicit evidence-to-equation-to-mechanism case object. \\
\midrule
MathVista \citep{lu2024mathvista}; MMMU \citep{yue2024mmmu}; SPIQA \citep{pramanick2024spiqa} & Multimodal questions and paper figures &
Visual reasoning and scientific document QA &
Physical models and biological mechanisms are not first-class grounding targets. \\
\midrule
PDEBench \citep{takamoto2022pdebench}; BubbleML \citep{jakli2023bubbleml} & Simulation data and physical benchmark tasks &
Physics-grounded prediction with controlled ground truth &
Focuses on numerical modeling rather than literature-grounded mechanism decisions. \\
\midrule
BioPhys-Bridge (ours) & Evidence-linked scientific reasoning cases &
Evidence IDs, quantitative values, equations, assumptions, mechanisms, and next decisions &
Designed for evidence-grounded equation-to-mechanism reasoning over biophysical literature. \\
\bottomrule
\end{tabular}
\caption{Positioning BioPhys-Bridge against recent benchmark families. The
distinguishing unit is not a question alone, but a structured case linking
source evidence, physical model, quantitative evidence, biological mechanism,
and scientific decision.}
\label{tab:benchmark-positioning}
\end{table*}

\section{Dataset Design}

\subsection{Evidence-Grounded Scientific Reasoning Cases}

\begin{figure*}[t]
\centering
\includegraphics[width=0.7\textwidth]{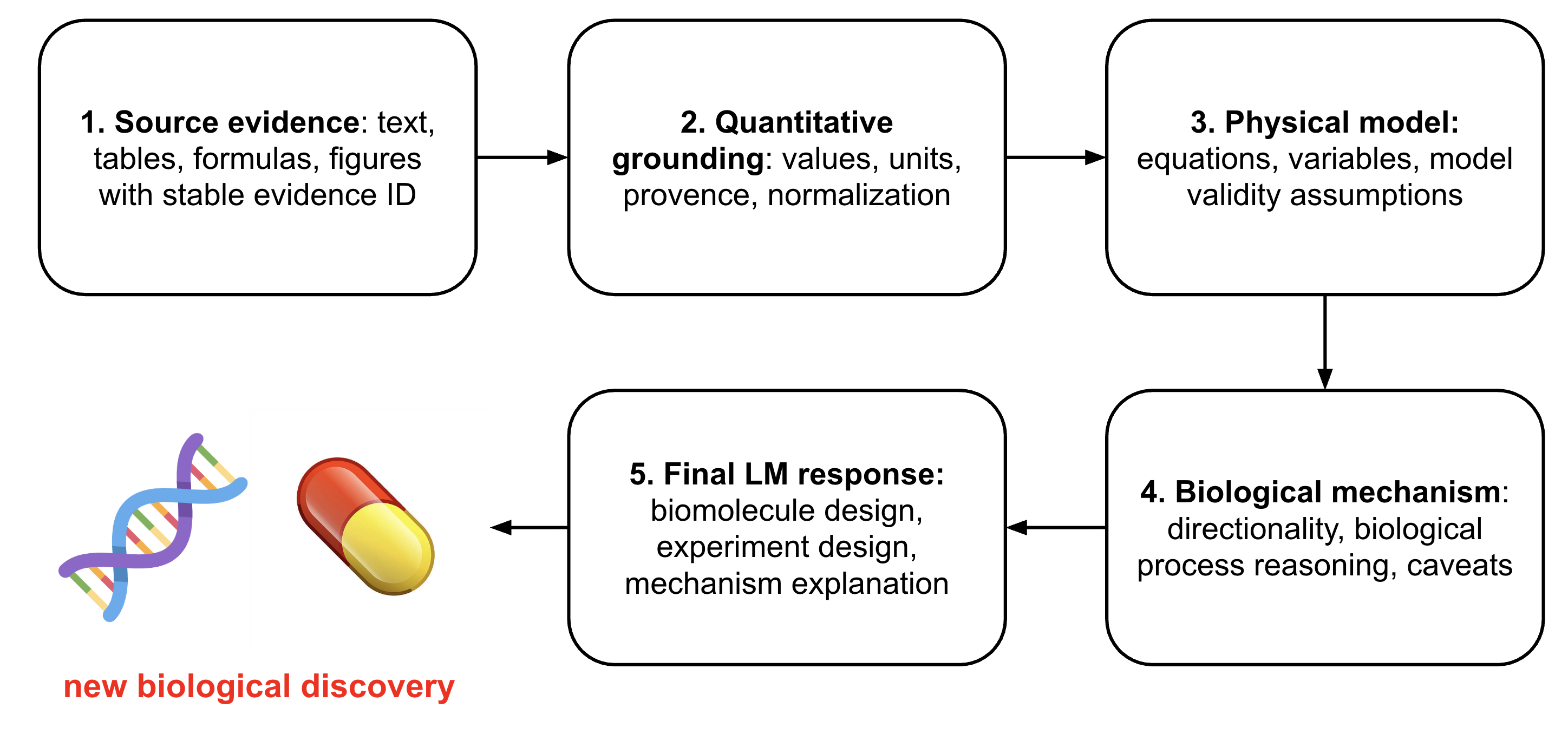}
\caption{Each case as an evidence-linked bridge from
source-paper evidence to quantitative values, physical-model assumptions,
mechanistic interpretation, and an agent-facing scientific decision task.}
\label{fig:case-structure}
\end{figure*}

A BioPhys-Bridge record is a structured grounding object rather than a
standalone QA item. The core reasoning pattern is:
\[
\begin{aligned}
\text{evidence} &\rightarrow \text{quantitative value}
\rightarrow \text{physical model}\\
&\rightarrow \text{mechanism} \rightarrow \text{next decision}.
\end{aligned}
\]
Figure~\ref{fig:case-structure} summarizes this structure. The schema
separates the biological setting from the physical model family. The field
\texttt{domain} captures the scientific area, while
\texttt{biophysical\_model.model\_family} records the primary physical
modeling strategy used by the main equation and decision. Evidence blocks are
the atomic citation units: all supporting evidence IDs in tasks must resolve
to these blocks, and quantitative evidence must be traceable to cited evidence
text.

Appendix~\ref{app:gold-case} gives a concrete expert-annotated gold-case
schematic for the same evidence-to-decision structure.

\subsection{Record Fields}

Each case includes paper provenance, evidence blocks, quantitative evidence,
model annotation, mechanism annotation, a staged trajectory, agent tasks, and
quality metadata. Table~\ref{tab:schema} gives the main fields. The
source-of-truth schema is implemented in Pydantic and exported as JSON Schema;
unknown fields are forbidden and cross-field invariants are checked during
validation. The expected model output for each task includes both an answer
and supporting evidence IDs, making evidence attribution part of the task
rather than post-hoc explanation.

\begin{table*}[t]
\centering
\small
\begin{tabular}{ll}
\toprule
Field & Purpose \\
\midrule
\texttt{case\_id} & Stable identifier for an evidence-grounded case. \\
\texttt{domain}, \texttt{bridge\_type} & Biological setting and domain-specific physics-to-mechanism bridge. \\
\texttt{source} & DOI/PMCID, title, license, source URL, and MinerU parse identifier. \\
\texttt{evidence[]} & Text, table, formula, and figure/caption blocks with stable evidence IDs. \\
\texttt{quantitative\_evidence[]} & Metric/value/unit records citing evidence IDs. \\
\texttt{biophysical\_model} & Model name, family, equation, variables, assumptions, validity conditions. \\
\texttt{physical\_interpretation} & Derived quantity, directionality, consistency notes, and caveats. \\
\texttt{biological\_mechanism} & Mechanism type and text linking physical result to biology. \\
\texttt{sci\_evo\_trajectory[]} & Stages from research question through observation, interpretation, and next step. \\
\texttt{agent\_tasks[]} & Evaluation tasks with gold answers and supporting evidence IDs. \\
\texttt{expert\_annotation} & Expert review notes when present. \\
\texttt{quality} & Validation status, review status, score, provenance, and audit metadata. \\
\bottomrule
\end{tabular}
\caption{Dataset schema of BioPhys-Bridge. Evidence
IDs, quantitative values, equations, assumptions, mechanisms, and decisions
are explicit grounding targets.}
\label{tab:schema}
\end{table*}

\subsection{Task Types}

The release contains four task types, each designed to test a different
grounding challenge. \texttt{derivation} tasks ask a model to use a physical
relation or equation while citing the relevant evidence IDs.
\texttt{mechanism\_from\_evidence} tasks require a biological mechanism
inferred from cited quantitative or textual evidence.
\texttt{discrepancy\_explanation} tasks require an explanation of tension
between expected physical behavior and observed evidence, grounded in the
case's evidence blocks. \texttt{next\_experiment\_design} tasks require a
follow-up experiment or computation grounded in cited evidence and model
assumptions.

\section{Data Curation Pipeline}

The data curation pipeline converts scientific research papers into grounding
benchmark records. It is designed to be traceable, resumable, and compliant
with source licensing. Figure~\ref{fig:pipeline} sketches the main stages; the
archived release audit funnel is reported in Appendix~\ref{app:audit-funnel}.

\begin{figure*}[t]
\centering
\includegraphics[width=0.7\textwidth]{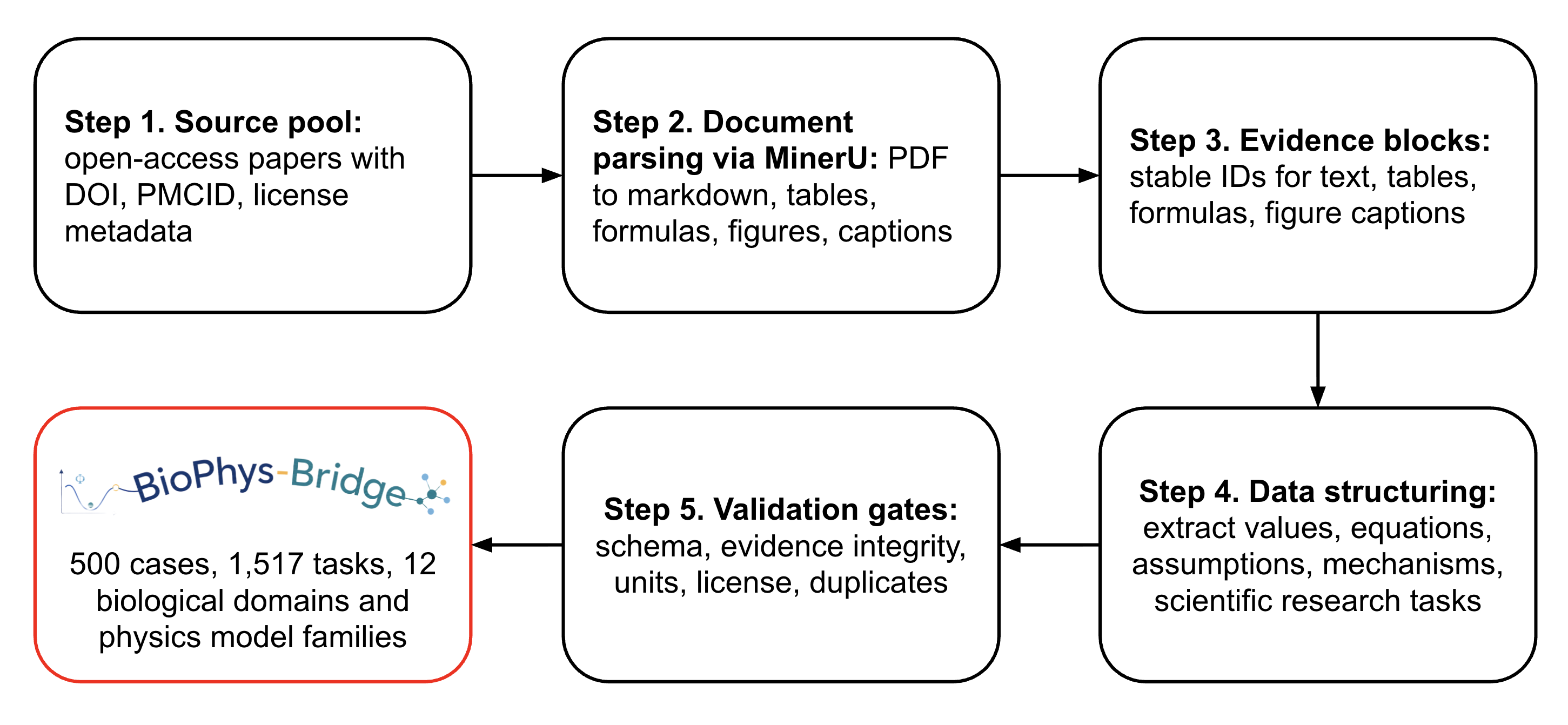}
\caption{Curation pipeline. Open-access papers are parsed with MinerU,
normalized into evidence blocks, structured into case records, and filtered by
validation and quality gates before release.}
\label{fig:pipeline}
\end{figure*}

\paragraph{Source acquisition and batching.}
Candidate papers are drawn from open-access scientific sources with DOI and/or
PMCID provenance and release-compatible licenses. Candidate metadata includes a
title, source URL, license, abstract, domain guess, model-family guess, and
model or quantitative keywords. Papers are selected in batches while tracking
coverage across domains and physical model families.

\paragraph{Document parsing and evidence block construction.}
Each PDF is parsed with MinerU. The raw PDF files and MinerU intermediate
payloads are retained outside the public release, while normalized evidence
artifacts are used to build release records. Normalization converts parsed
outputs into canonical document folders with markdown, structured JSON,
evidence blocks, and parse metadata. Evidence blocks are preserved as the
atomic citation units used by model prompts and evidence-ID scoring.

\paragraph{Grounded candidate extraction and structuring.}
A regex-first pass extracts candidate numeric values, equations, units, model
keywords, and evidence-rich snippets from prose, tables, formulas, and figure
captions. An evidence-only LLM pass may then structure fields such as
quantitative evidence, physical interpretation, mechanism text, and agent
tasks. The shipped release used OpenAI \texttt{gpt-4o} for this evidence-only
structuring pass. The prompt instructs the model to use only supplied evidence
and to leave unsupported fields empty rather than hallucinate measurements,
equations, assumptions, or mechanisms. Unsupported or fabricated evidence IDs
are removed by release validation.

\paragraph{Release export.}
Final export writes JSONL records, deterministic splits, metadata, schema, a
dataset card, gold samples, and a Sci-Evo view. Source PDFs, raw MinerU
payloads, and LLM responses are not redistributed.

\paragraph{Expert annotation layer.}
Expert annotation is tracked separately from release-gate review. The full
held-out test split, 10 contest gold samples, 30 extended-gold samples, and one
legacy reviewed record include physics reasoning, biological reasoning,
uncertainty, and reviewer notes. This separation prevents the release-gate
field \texttt{quality.manual\_review\_status} from being mistaken for a claim
that all 500 cases have full expert biological and physical review.

\section{Release Statistics}

Table~\ref{tab:overview} summarizes the current release. The dataset contains
500 cases and 1,517 agent tasks. The split is deterministic and stratified by
domain, with 400 training cases, 50 validation cases, and 50 test cases. It is
also grouped by source paper under DOI/PMCID/title keys: the train, validation,
and test splits share zero source papers. The release includes 107 cases with
an explicit failure or revision stage, and 81 cases with expert annotation. The
expert annotation count covers the 50 held-out test cases, 10 contest gold
samples, 30 extended-gold samples, and one legacy reviewed record.

\begin{table}[t]
\centering
\small
\begin{tabular}{lr}
\toprule
Release item & Value \\
\midrule
Cases & 500 \\
Train / validation / test & 400 / 50 / 50 \\
Agent tasks & 1,517 \\
Failure/revision cases & 107 \\
Expert-annotation cases & 81 \\
Mean modalities per case & 2.944 \\
Cases with 3+ modalities & 80.4\% \\
Schema-valid rate & 100\% \\
Evidence coverage rate & 100\% \\
Quantitative-evidence rate & 100\% \\
Source-license coverage & 100\% \\
Duplicate rate & 0\% \\
\bottomrule
\end{tabular}
\caption{BioPhys-Bridge v1 release at a glance.}
\label{tab:overview}
\end{table}

Table~\ref{tab:domains} shows the weighted domain coverage. The release is not
intended to be perfectly balanced. Instead, it emphasizes coverage across
biophysical discovery settings while preserving enough examples in the larger
protein-ligand and systems-biology groups to support model training and
evaluation.

\begin{table}[t]
\centering
\small
\begin{tabular}{lr}
\toprule
Biological domain & Cases \\
\midrule
\texttt{protein\_ligand\_binding} & 188 \\
\texttt{systems\_biology\_dynamics} & 126 \\
\texttt{conformational\_dynamics\_allostery} & 91 \\
\texttt{biomolecular\_phase\_separation} & 52 \\
\texttt{enzyme\_kinetics} & 22 \\
\texttt{protein\_stability\_thermodynamics} & 21 \\
\bottomrule
\end{tabular}
\caption{Biological-domain coverage in the 500-case release.}
\label{tab:domains}
\end{table}

\begin{table}[t]
\centering
\small
\begin{tabular}{lr}
\toprule
Task type & Count \\
\midrule
\texttt{mechanism\_from\_evidence} & 480 \\
\texttt{next\_experiment\_design} & 403 \\
\texttt{derivation} & 355 \\
\texttt{discrepancy\_explanation} & 279 \\
\bottomrule
\end{tabular}
\caption{Agent-task distribution.}
\label{tab:tasks}
\end{table}

The primary physical model families include binding thermodynamics (185),
systems stochastic dynamics (128), conformational/allosteric energy landscapes
(93), polymer phase-separation statistical mechanics (41), enzyme reaction
kinetics (23), folding stability thermodynamics (20), spatial transport and
electrostatics (6), evolutionary fitness landscapes (2), and mechanical force
response (2). The last three families are included for coverage and future
expansion, not for standalone per-family evaluation in this release.

\section{Quality Control and Validation}

Quality control is central to the release. Table~\ref{tab:quality} summarizes
the main gates. A shipped case must validate against the schema, cite valid
evidence IDs, ground quantitative values in cited evidence text, carry source
license metadata, pass duplicate checks, and avoid unresolved templates or weak
task prompts. Unit normalization retains both raw and normalized units when
normalization is supported. The content gate blocks examples with unsupported
measurements, fabricated evidence IDs, evidence-ID-only prompts, unresolved
template strings, malformed tool vocabularies, and missing next-step stages.

\begin{table*}[t]
\centering
\small
\begin{tabular}{lll}
\toprule
Gate & Check & Release result \\
\midrule
Schema validation & Pydantic and generated JSON Schema & 500 / 500 valid \\
Evidence integrity & Cited IDs resolve to evidence blocks & 100\% coverage \\
Quantitative grounding & Values appear in cited evidence text & 100\% coverage \\
Source compliance & Release-compatible source licenses & 490 CC-BY-4.0, 10 CC0-1.0 \\
Unit normalization & Raw and normalized units retained & 100\% success \\
Duplicate check & Stable case IDs are unique & 0 duplicates \\
Content gate & No unresolved templates or weak tasks & 100\% pass \\
Expert annotation & Selected gold and test records & 81 cases \\
Audit notes & Deterministic notes for shipped cases & 100\% note coverage \\
\bottomrule
\end{tabular}
\caption{Release quality gates. The deterministic audit notes are a supporting
traceability aid; only
about 2\% of cases have an implemented closed-form relation-level check, so it
is not used as a global correctness metric.}
\label{tab:quality}
\end{table*}

The field \texttt{quality.manual\_review\_status = reviewed} should be read as
a release-gate status, not as a claim that all 500 cases have full expert
physics and biology annotation. Full expert annotations are tracked separately
through the \texttt{expert\_annotation} field.

\section{Evaluation Protocol and Baselines}

\paragraph{Evaluation setting.}
Given a task prompt and candidate evidence blocks, the model must produce a
JSON answer and a set of supporting evidence IDs. The scorer computes
evidence-ID F1 against the gold supporting evidence IDs and answer token F1
against the gold answer. Evidence-ID F1 evaluates whether the model grounds
its answer in the correct evidence blocks. Answer token F1 is a rough
sanity-check metric for this initial release, not a final measure of
scientific correctness. The released harness also computes
\texttt{overall\_score} as $0.7$ answer token F1 plus $0.3$ evidence-ID F1 for
diagnostics, but we do not use that composite metric for the main paper
because token F1 is weak for open-ended scientific answers.

\paragraph{De-leaked prompt construction.}
The evaluation prompt hides gold answers, gold supporting evidence IDs, and
expert annotations. The main de-leaked prompt exposes the task domain, task
type, question, and ranked candidate evidence blocks, but not the structured
equation, physical directionality, or biological mechanism fields. We use this
no-scaffold prompt for the main results because those structured fields can be
answer-bearing for derivation or mechanism tasks. Candidate evidence selection
uses a lexical ranker over only the task type, task question, research
question, domain, and source evidence text; it does not access
\texttt{task.supporting\_evidence\_ids} or structured target annotations such
as equations, physical directionality, biological mechanisms, or quantitative
value records. Gold IDs are used only by the scorer after prediction. In the
released harness, prompt construction and candidate selection operate on
public case/task views that exclude \texttt{gold\_answer},
\texttt{supporting\_evidence\_ids}, and \texttt{expert\_annotation};
regression tests fail if these fields reach the model-visible evaluation path.
Each test task is shown 48 candidate evidence blocks selected from a median
206 evidence blocks per case; the deterministic lexical baseline cites the top
3 ranked blocks. This candidate generator contains 234 of 267 gold supporting
evidence IDs on the test set (0.876 gold-ID recall), with all gold IDs present
for 127 of 154 tasks; the mean maximum achievable evidence-ID F1 under the
candidate set is 0.916. Thus the
reported scores evaluate attribution and re-ranking within a lexically
pre-filtered candidate set, not unconstrained retrieval over every evidence
block. Regression tests verify that prompts and the lexical baseline do not
copy gold evidence IDs. All LLM rows use temperature 0 and JSON-mode decoding.
Confidence intervals are nonparametric bootstrap 95\% intervals over tasks.

\begin{table*}[t]
\centering
\scriptsize
\begin{tabular}{llrrr}
\toprule
Model & Provider & Evidence-ID F1 & Answer token F1 & Empty EID outputs \\
\midrule
Lexical retrieval & deterministic & 0.188 [0.151, 0.226] & 0.058 [0.053, 0.065] & 0/154 \\
DeepSeek v4 Flash & DeepSeek & 0.360 [0.309, 0.412] & 0.193 [0.173, 0.214] & 0/154 \\
Qwen 3.7 Max & OpenRouter & 0.316 [0.273, 0.359] & 0.129 [0.115, 0.143] & 1/154 \\
GPT-4o-mini & OpenRouter & 0.294 [0.251, 0.339] & 0.163 [0.149, 0.177] & 0/154 \\
Claude Opus 4.8 & OpenRouter & 0.238 [0.205, 0.272] & 0.095 [0.083, 0.107] & 18/154 \\
Claude Sonnet 5 & OpenRouter & 0.224 [0.190, 0.258] & 0.077 [0.068, 0.086] & 21/154 \\
DeepSeek v4 Pro & DeepSeek & 0.111 [0.070, 0.156] & 0.046 [0.029, 0.065] & 127/154 \\
Gemini 2.5 Pro & OpenRouter & 0.010 [0.000, 0.023] & 0.099 [0.084, 0.115] & 151/154 \\
\bottomrule
\end{tabular}
\caption{Preliminary de-leaked no-scaffold evaluation on the identical
154-task held-out test set. Values in brackets are bootstrap 95\% confidence
intervals over tasks. Empty EID outputs counts predictions with no parsed
\texttt{supporting\_evidence\_ids}; this is an output-compliance measure as
well as an attribution failure mode.}
\label{tab:eval}
\end{table*}

Paired bootstrap over the same 154 tasks gives evidence-ID F1 deltas over
lexical retrieval of $+0.172$ for DeepSeek v4 Flash (95\% CI
$[0.117, 0.227]$), $+0.127$ for Qwen 3.7 Max (95\% CI $[0.076, 0.179]$), and
$+0.106$ for GPT-4o-mini (95\% CI $[0.061, 0.152]$). Claude Opus 4.8 shows a
smaller positive delta ($+0.050$, 95\% CI $[0.005, 0.094]$), while Claude
Sonnet 5 is not significantly above lexical retrieval under this paired
bootstrap ($+0.035$, 95\% CI $[-0.010, 0.081]$). DeepSeek v4 Flash also
outperforms Qwen 3.7 Max by $+0.045$ evidence-ID F1 (95\% CI
$[0.012, 0.079]$). We retain a scaffolded prompt mode for diagnostic audits,
but exclude it from the main baseline because structured fields can be
answer-bearing, including equations that match derivation gold answers.

\begin{table*}[t]
\centering
\scriptsize
\begin{tabular}{lrrrrr}
\toprule
Task type & N & Lex. & Flash & Qwen & GPT-mini \\
\midrule
\texttt{derivation} & 39 & 0.192 & 0.475 & 0.380 & 0.346 \\
\texttt{discrepancy} & 31 & 0.201 & 0.404 & 0.379 & 0.350 \\
\texttt{mechanism} & 46 & 0.188 & 0.352 & 0.311 & 0.306 \\
\texttt{next experiment} & 38 & 0.173 & 0.216 & 0.203 & 0.181 \\
\bottomrule
\end{tabular}
\caption{Held-out evidence-ID F1 by task type for the lexical floor and the
three strongest LLM rows in Table~\ref{tab:eval}. Each task-type cell contains
only 31--46 tasks, so these point estimates should be read as noisier
diagnostics rather than stable per-type rankings.}
\label{tab:task-breakdown}
\end{table*}

These baselines are intentionally preliminary. The same-test comparison shows
that several LLMs improve over lexical retrieval in aggregate evidence
attribution, with the largest gains on \texttt{derivation},
\texttt{discrepancy\_explanation}, and \texttt{mechanism\_from\_evidence}
tasks. \texttt{next\_experiment\_design} remains harder: even the strongest
model reaches only 0.216 evidence-ID F1, consistent with the task's open-ended
decision structure. The panel also exposes a grounded-generation failure mode:
some models produce plausible prose but fail to emit usable evidence IDs.
Gemini 2.5 Pro has nontrivial answer-token overlap (0.099) but empty parsed
evidence IDs for 151 of 154 tasks, making its attribution score near zero.
DeepSeek v4 Pro similarly leaves 127 of 154 outputs without parsed evidence
IDs. Answer token F1 therefore remains a weak proxy for grounded scientific
answer quality, while evidence-ID F1 captures whether the answer is actually
attributable to the supplied evidence blocks.

Because scientific answers are open-ended, future releases should add
rubric-based expert scoring for physical-model use, quantitative consistency,
mechanism correctness, uncertainty handling, and experimental feasibility.
This is especially important for grounded scientific agents, where a cited but
mechanistically wrong answer can still receive partial evidence overlap.
The 50 test cases have expert annotation notes, but they are not independent
parallel task-level evidence-ID labels, so we do not report inter-annotator
agreement for supporting evidence IDs in this release.

\section{Dataset Access and Reproducibility}

For peer review, the dataset, schema, aggregate reports, and evaluation
outputs are supplied in the supplementary materials. The public dataset and
code repository will be released after peer review. The repository will
contain code, schema, reports, validation tests, small samples, aggregate
summaries, and release files. Code will be released under the MIT License,
while curated dataset metadata, reports, and public release files will be
released under CC-BY-4.0. Upstream source papers retain their original
licenses, recorded per case. Raw PDFs, raw MinerU payloads, and LLM responses
are not redistributed.

The public release will expose validation and evaluation commands through the
\texttt{biophysevo} Python package. The source-of-truth schema lives in
\texttt{src/biophysevo/schemas/case\_schema.py}, and the generated JSON Schema
is included in the supplementary materials.

\section{Limitations and Responsible Use}

BioPhys-Bridge is an initial benchmark release, not a substitute for reading
the original papers and not a complete solution to biological reasoning. Users
should verify important claims against the cited source literature before
using them in research, engineering, clinical, or safety-critical settings.
Expert annotation covers 81 cases, not all 500. Current baselines are
preliminary: under the fully de-leaked public-query setting, they include a
lexical retrieval floor and seven LLMs, but not a human ceiling, hidden test
server, repeated stochastic runs, or full rubric-based expert grading.
Evidence-ID F1 captures attribution to evidence blocks but not full scientific
correctness, and token F1 is a weak
proxy for open-ended answers. The expert layer checks physical and biological
reasoning for the test cases, but it does not provide independent parallel
task-level labels for supporting evidence IDs; future versions should add
inter-annotator agreement and expert scoring for mechanism correctness.
Because the release structuring pass used GPT-4o, shared-provider bias remains
a caveat for the GPT-4o-mini row, although the strongest reported evidence-ID
scores come from non-OpenAI systems. Model routing and output-format behavior
also affect measured attribution: systems that omit parseable
\texttt{supporting\_evidence\_ids} can receive very low evidence-ID F1 even
when their prose answer overlaps the reference answer.

The test set is public in this release, which supports reuse and workshop
discussion but limits long-term contamination control; future versions should
include hidden or contamination-aware evaluation. The source pool is
open-access and release-compatible, which induces domain and source bias.
Domain coverage is weighted rather than balanced. OCR and table parsing
artifacts can remain in evidence text, because preserving traceability to
parsed source artifacts is prioritized over silently rewriting source
evidence. The archived release metadata begins at the reviewed candidate set
rather than the full exploratory source pool, so future versions should report
complete rejection and parser-failure counts.

\section{Conclusion}

We introduce BioPhys-Bridge, an initial benchmark for evidence-grounded
scientific reasoning over biophysical literature. The current 500-case release
is reusable and validated enough for community exploration: it represents
evidence blocks, quantitative values, equations, assumptions, mechanisms, and
next scientific decisions in a shared case schema, and it provides de-leaked
baselines for evidence-ID attribution. Preliminary same-test results show that
several LLMs improve over lexical retrieval for evidence attribution, with
DeepSeek v4 Flash, Qwen 3.7 Max, and GPT-4o-mini producing the strongest
evidence-ID F1 in our initial panel. The results also show that grounded
answering is not only answer generation: models can produce plausible prose
while failing to emit usable evidence IDs. This
workshop version is intended to invite feedback on evaluation design,
expert-rubric scoring, contamination-aware protocols, and future grounded
scientific-agent benchmarks that test not only what a paper says, but how
source evidence supports quantitative and mechanistic scientific decisions.

\section*{Acknowledgments}

Acknowledgments are omitted for anonymous review.

\section*{Declaration on Generative AI}

During the preparation of this work, the author used ChatGPT for brainstorming,
paper drafting and editing, and code scaffolding, and used Claude Code for code
implementation. The author reviewed and edited the entire content and takes
full responsibility for the publication's content.

\bibliography{references}

@inproceedings{lu2022scienceqa,
  title = {Learn to Explain: Multimodal Reasoning via Thought Chains for Science Question Answering},
  author = {Lu, Pan and Mishra, Swaroop and Xia, Tony and Qiu, Liang and Chang, Kai-Wei and Zhu, Song-Chun and Tafjord, Oyvind and Clark, Peter and Kalyan, Ashwin},
  booktitle = {Advances in Neural Information Processing Systems},
  volume = {35},
  pages = {2507--2521},
  year = {2022},
  eprint = {2209.09513},
  archivePrefix = {arXiv},
  primaryClass = {cs.CL}
}

@inproceedings{wang2023scibench,
  title = {SciBench: Evaluating College-Level Scientific Problem-Solving Abilities of Large Language Models},
  author = {Wang, Xiaoxuan and Hu, Ziniu and Lu, Pan and Zhu, Yanqiao and Zhang, Jieyu and Subramaniam, Satyen and Loomba, Arjun R. and Zhang, Shichang and Sun, Yizhou and Wang, Wei},
  booktitle = {Proceedings of the 41st International Conference on Machine Learning},
  publisher = {PMLR},
  pages = {50622--50649},
  year = {2024},
  eprint = {2307.10635},
  archivePrefix = {arXiv},
  primaryClass = {cs.CL}
}

@inproceedings{rein2023gpqa,
  title = {GPQA: A Graduate-Level Google-Proof Q\&A Benchmark},
  author = {Rein, David and Hou, Betty Li and Stickland, Asa Cooper and Petty, Jackson and Pang, Richard Yuanzhe and Dirani, Julien and Michael, Julian and Bowman, Samuel R.},
  booktitle = {First Conference on Language Modeling},
  year = {2024},
  url = {https://openreview.net/forum?id=Ti67584b98},
  eprint = {2311.12022},
  archivePrefix = {arXiv},
  primaryClass = {cs.AI}
}

@misc{laurent2024labbench,
  title = {LAB-Bench: Measuring Capabilities of Language Models for Biology Research},
  author = {Laurent, Jon M. and Janizek, Joseph D. and Ruzo, Michael and Hinks, Michaela M. and Hammerling, Michael J. and Narayanan, Siddharth and Ponnapati, Manvitha and White, Andrew D. and Rodriques, Samuel G.},
  year = {2024},
  eprint = {2407.10362},
  archivePrefix = {arXiv},
  primaryClass = {cs.CL}
}

@misc{lala2023paperqa,
  title = {PaperQA: Retrieval-Augmented Generative Agent for Scientific Research},
  author = {Lala, Jakub and O'Donoghue, Odhran and Shtedritski, Aleksandar and Cox, Sam and Rodriques, Samuel G. and White, Andrew D.},
  year = {2023},
  eprint = {2312.07559},
  archivePrefix = {arXiv},
  primaryClass = {cs.CL}
}

@misc{wang2024mineru,
  title = {MinerU: An Open-Source Solution for Precise Document Content Extraction},
  author = {Wang, Bin and Xu, Chao and Zhao, Xiaomeng and Ouyang, Linke and Wu, Fan and Zhao, Zhiyuan and Xu, Rui and Liu, Kaiwen and Qu, Yuan and Shang, Fukai and others},
  year = {2024},
  eprint = {2409.18839},
  archivePrefix = {arXiv},
  primaryClass = {cs.CV}
}

@inproceedings{ajith2024litsearch,
  title = {LitSearch: A Retrieval Benchmark for Scientific Literature Search},
  author = {Ajith, Anirudh and Xia, Mengzhou and Chevalier, Alexis and Goyal, Tanya and Chen, Danqi and Gao, Tianyu},
  booktitle = {Proceedings of the 2024 Conference on Empirical Methods in Natural Language Processing},
  address = {Miami, Florida, USA},
  publisher = {Association for Computational Linguistics},
  year = {2024},
  pages = {15068--15083},
  url = {https://aclanthology.org/2024.emnlp-main.840/},
  doi = {10.18653/v1/2024.emnlp-main.840}
}

@inproceedings{lu2024mathvista,
  title = {MathVista: Evaluating Mathematical Reasoning of Foundation Models in Visual Contexts},
  author = {Lu, Pan and Bansal, Hritik and Xia, Tony and Liu, Jiacheng and Li, Chunyuan and Hajishirzi, Hannaneh and Cheng, Hao and Chang, Kai-Wei and Galley, Michel and Gao, Jianfeng},
  booktitle = {International Conference on Learning Representations},
  year = {2024}
}

@inproceedings{yue2024mmmu,
  title = {MMMU: A Massive Multi-discipline Multimodal Understanding and Reasoning Benchmark for Expert AGI},
  author = {Yue, Xiang and Ni, Yuansheng and Zhang, Kai and Zheng, Tianyu and Liu, Ruoqi and Zhang, Ge and Stevens, Samuel and Jiang, Dongfu and Ren, Weiming and Sun, Yuxuan and others},
  booktitle = {Proceedings of the IEEE/CVF Conference on Computer Vision and Pattern Recognition},
  year = {2024}
}

@inproceedings{bai2024longbench,
  title = {LongBench: A Bilingual, Multitask Benchmark for Long Context Understanding},
  author = {Bai, Yushi and Lv, Xin and Zhang, Jiajie and Lyu, Hongchang and Tang, Jiankai and Huang, Zhidian and Du, Zhengxiao and Liu, Xiao and Zeng, Aohan and Hou, Lei and Dong, Yuxiao and Tang, Jie and Li, Juanzi},
  booktitle = {Proceedings of the 62nd Annual Meeting of the Association for Computational Linguistics (Volume 1: Long Papers)},
  pages = {3119--3137},
  address = {Bangkok, Thailand},
  publisher = {Association for Computational Linguistics},
  year = {2024},
  url = {https://aclanthology.org/2024.acl-long.172/},
  doi = {10.18653/v1/2024.acl-long.172}
}

@inproceedings{pramanick2024spiqa,
  title = {SPIQA: A Dataset for Multimodal Question Answering on Scientific Papers},
  author = {Pramanick, Shraman and Chellappa, Rama and Venugopalan, Subhashini},
  booktitle = {Advances in Neural Information Processing Systems, Datasets and Benchmarks Track},
  year = {2024}
}

@inproceedings{takamoto2022pdebench,
  title = {PDEBench: An Extensive Benchmark for Scientific Machine Learning},
  author = {Takamoto, Makoto and Praditia, Timothy and Leiteritz, Raphael and MacKinlay, Dan and Alesiani, Francesco and Pfl{\"u}ger, Dirk and Niepert, Mathias},
  booktitle = {Advances in Neural Information Processing Systems, Datasets and Benchmarks Track},
  year = {2022}
}

@inproceedings{jakli2023bubbleml,
  title = {BubbleML: A Multiphase Multiphysics Dataset and Benchmarks for Machine Learning},
  author = {Hassan, Sheikh Md Shakeel and Feeney, Arthur and Dhruv, Akash and Kim, Jihoon and Suh, Youngjoon and Ryu, Jaiyoung and Won, Yoonjin and Chandramowlishwaran, Aparna},
  booktitle = {Advances in Neural Information Processing Systems, Datasets and Benchmarks Track},
  year = {2023}
}

@inproceedings{jimenez2024swebench,
  title = {SWE-bench: Can Language Models Resolve Real-World GitHub Issues?},
  author = {Jimenez, Carlos E. and Yang, John and Wettig, Alexander and Yao, Shunyu and Pei, Kexin and Press, Ofir and Narasimhan, Karthik},
  booktitle = {International Conference on Learning Representations},
  year = {2024}
}

@inproceedings{zhou2024webarena,
  title = {WebArena: A Realistic Web Environment for Building Autonomous Agents},
  author = {Zhou, Shuyan and Xu, Frank F. and Zhu, Hao and Zhou, Xuhui and Lo, Robert and Sridhar, Abishek and Cheng, Xianyi and Ou, Tianyue and Bisk, Yonatan and Fried, Daniel and Alon, Uri and Neubig, Graham},
  booktitle = {International Conference on Learning Representations},
  year = {2024}
}

@inproceedings{liu2024agentbench,
  title = {AgentBench: Evaluating LLMs as Agents},
  author = {Liu, Xiao and Yu, Hao and Zhang, Hanchen and Xu, Yifan and Lei, Xuanyu and Lai, Hanyu and Gu, Yu and Ding, Hangliang and Men, Kaiwen and Yang, Kejuan and others},
  booktitle = {International Conference on Learning Representations},
  year = {2024}
}

@inproceedings{wang2024mint,
  title = {MINT: Evaluating LLMs in Multi-turn Interaction with Tools and Language Feedback},
  author = {Wang, Xingyao and Wang, Zihan and Liu, Jiateng and Chen, Yangyi and Yuan, Lifan and Peng, Hao and Ji, Heng},
  booktitle = {International Conference on Learning Representations},
  year = {2024}
}

@inproceedings{qin2024toolllm,
  title = {ToolLLM: Facilitating Large Language Models to Master 16000+ Real-world APIs},
  author = {Qin, Yujia and Liang, Shihao and Ye, Yining and Zhu, Kunlun and Yan, Lan and Lu, Yaxi and Lin, Yankai and Cong, Xin and Tang, Xiangru and Qian, Bill and Zhao, Sihan and Hong, Lauren and Tian, Runchu and Xie, Ruobing and Zhou, Jie and Gerstein, Mark and Li, Dahai and Liu, Zhiyuan and Sun, Maosong},
  booktitle = {International Conference on Learning Representations},
  year = {2024}
}

@inproceedings{shern2025mlebench,
  title = {MLE-bench: Evaluating Machine Learning Agents on Machine Learning Engineering},
  author = {Chan, Jun Shern and Chowdhury, Neil and Jaffe, Oliver and Aung, James and Sherburn, Dane and Mays, Evan and Starace, Giulio and Liu, Kevin and Maksin, Leon and Patwardhan, Tejal and Madry, Aleksander and Weng, Lilian},
  booktitle = {International Conference on Learning Representations},
  year = {2025}
}

\appendix

\section{Gold Case Example}
\label{app:gold-case}

\begin{figure*}[t]
\centering
\includegraphics[width=0.7\textwidth]{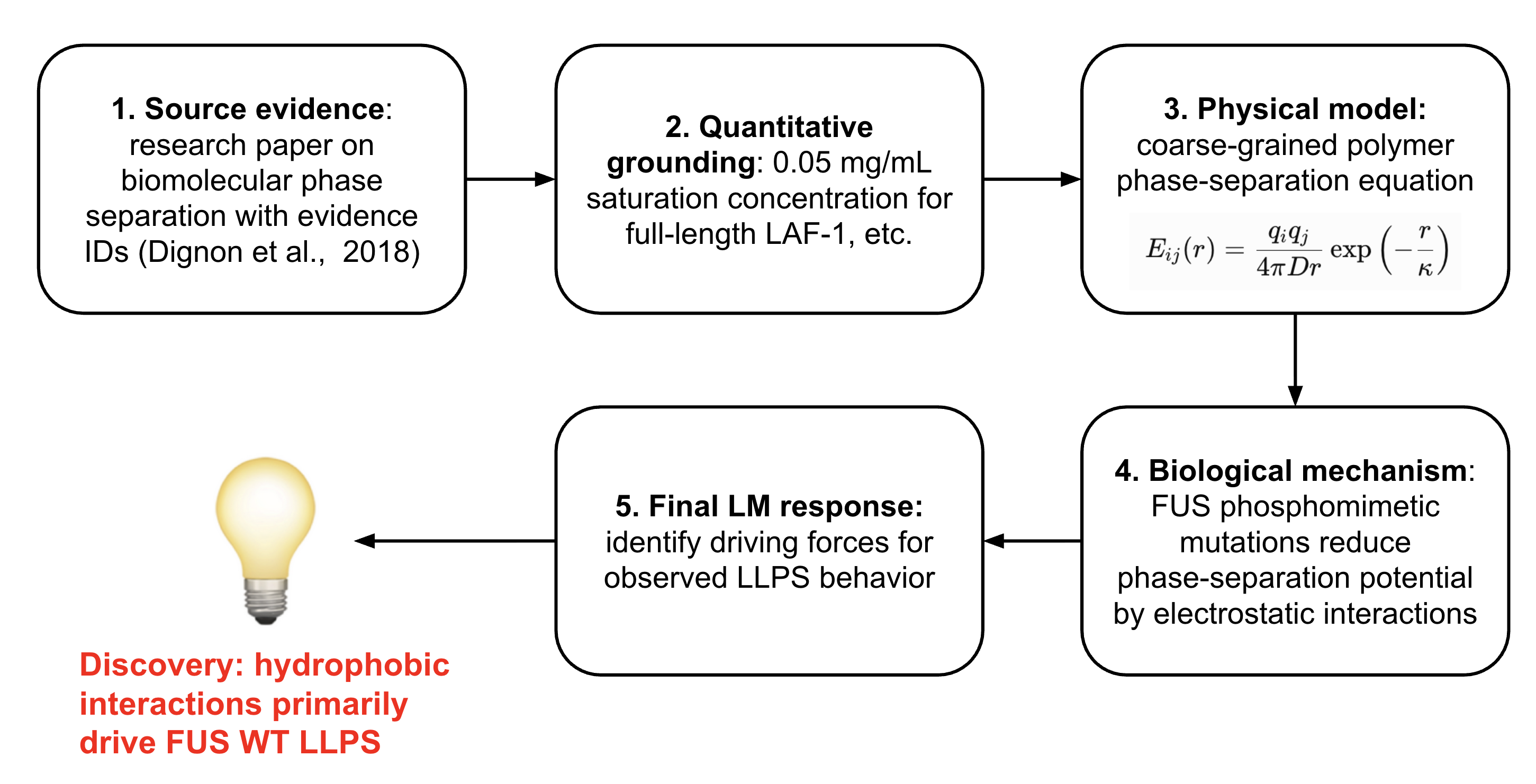}
\caption{A real BioPhys-Bridge case illustrates the benchmark object: source
provenance, evidence-linked quantitative measurements, a physical model,
mechanism interpretation, caveats, and an agent-facing task are stored together
rather than flattened into a standalone question.}
\label{fig:gold-case}
\end{figure*}

\section{Release Audit Funnel}
\label{app:audit-funnel}

\begin{table}[h]
\centering
\small
\begin{tabular}{lr}
\toprule
Archived release stage & Count \\
\midrule
Reviewed source records & 500 \\
Schema-valid source records & 500 \\
Exported case records & 500 \\
Deterministic train split & 400 \\
Deterministic validation split & 50 \\
Deterministic test split & 50 \\
Expert-annotated records & 81 \\
\bottomrule
\end{tabular}
\caption{Release audit funnel recorded in the public metadata. Future dataset
versions should additionally archive exploratory source-pool and parser-failure
counts to support a full rejection funnel.}
\label{tab:funnel}
\end{table}

\section{Prompt Template}

The evaluation prompt is generated by the released harness as follows:
\begin{quote}\small
\texttt{You are answering an evidence-grounded scientific question for biological research. Use only the evidence below. Return JSON with keys answer and supporting\_evidence\_ids. Choose supporting\_evidence\_ids from the candidate evidence IDs below; gold IDs are not provided. Domain: <domain>. Task type: <task\_type>. Evidence: <ranked evidence blocks>. Question: <task input>.}
\end{quote}
For evaluation, gold answers, gold supporting evidence IDs, and expert
annotations are hidden from the prompt; gold labels are used only by the
scorer.
The harness also retains a scaffolded diagnostic mode that adds
\texttt{Physical model}, \texttt{Equation}, \texttt{Physical directionality},
and \texttt{Biological mechanism} lines; it is not used for the main
de-leaked results because those fields can be answer-bearing.

\end{document}